\documentclass[letterpaper,10pt,conference]{ieeeconf}

\IEEEoverridecommandlockouts
\usepackage{amsmath}
\usepackage{amssymb}
\usepackage{booktabs}
\usepackage{graphicx}
\usepackage{microtype}
\usepackage{placeins}
\usepackage[table]{xcolor}
\usepackage{url}
\usepackage{tikz}
\usepackage{bm}
\usepackage{cite}
\usepackage{algorithm}
\usepackage{algpseudocode}
\usetikzlibrary{arrows.meta,backgrounds,fit,positioning,shapes.geometric}

\newcommand{\method}{RouteRLT}

\title{\LARGE \bf
Generalists Act, Specialists Intervene: Modular Stage-Selective Reinforcement Learning for Vision--Language--Action Manipulation
}

\author{Chongyu Zhu, Hyegang Kim, Jaden Hinds, Juan Sebastian Rojas,
Ramy Elmallah, Chi-Guhn Lee
\thanks{Department of Mechanical \& Industrial Engineering, University of Toronto.}%
\thanks{Correspondence: Chongyu Zhu, \texttt{cyzhu@mie.utoronto.ca}}}%

\usepackage[scaled=0.92]{helvet}
\definecolor{grouprow}{gray}{0.92}

\begin{document}

\maketitle
\thispagestyle{empty}
\pagestyle{empty}
\suppressfloats[t]
\begin{abstract}
Vision--language--action (VLA) models often struggle in the precision-critical phases of multi-stage manipulation tasks. To mitigate this issue, VLA models can be used in conjunction with reinforcement learning (RL) specialists that are specifically trained to handle the precision-critical phases. However, the \emph{coordination} between the base VLA model and the RL specialists, which dictates when a specialist should take over from the base VLA and vice-versa, remains an open research question. In this paper, we address this gap by introducing \emph{\method{}}, a modular framework that coordinates a generalist VLA, used as the default controller, with designated precision-critical RL specialists. At a high level, our framework trains a \emph{phase-aware coordination} mechanism that handles handoffs between the generalist and the specialists. We evaluate RouteRLT on the LIBERO and LIBERO-Plus benchmarks, as well as on a physical connector pickup and insertion task. Overall, we find that RouteRLT improves success on LIBERO, and retains net gains on LIBERO-Plus. On the physical task, RouteRLT completes 65.7\% of trials, compared with 8.6\% for the baseline. Altogether, these results demonstrate that learned coordination builds on generalist VLA capabilities to improve task completion in precision-critical manipulation.
\end{abstract}

\begin{figure}[t]
\centering
\includegraphics[width=\columnwidth]{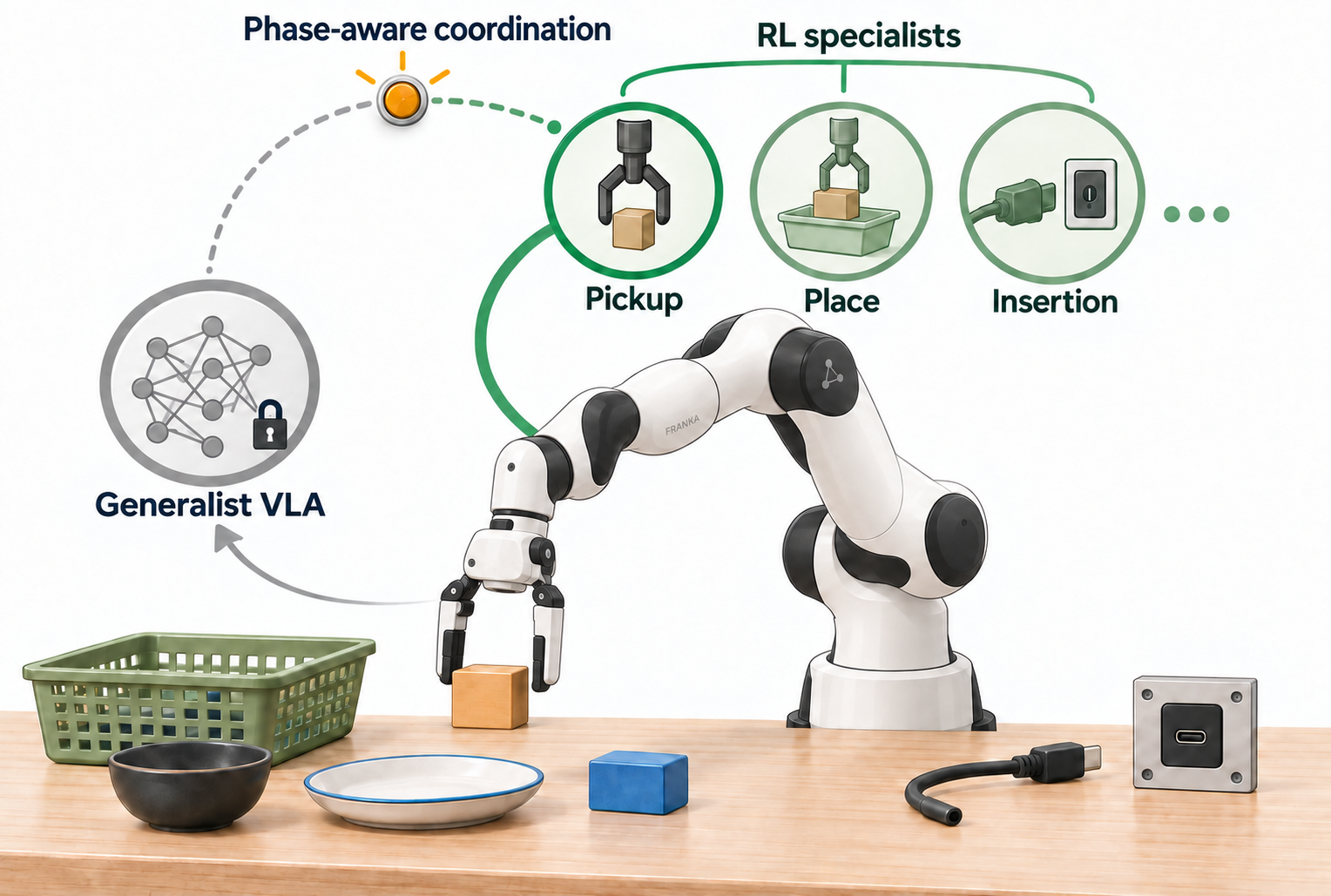}
\caption{Conceptual illustration of \method{}: the frozen VLA acts by default, while a phase-aware coordination mechanism selects the RL specialist and intervention timing for critical phases.}
\label{fig:routerlt_concept}
\vspace{-12pt}
\end{figure}

\begin{figure*}[!t]
\centering
\includegraphics[width=\textwidth]{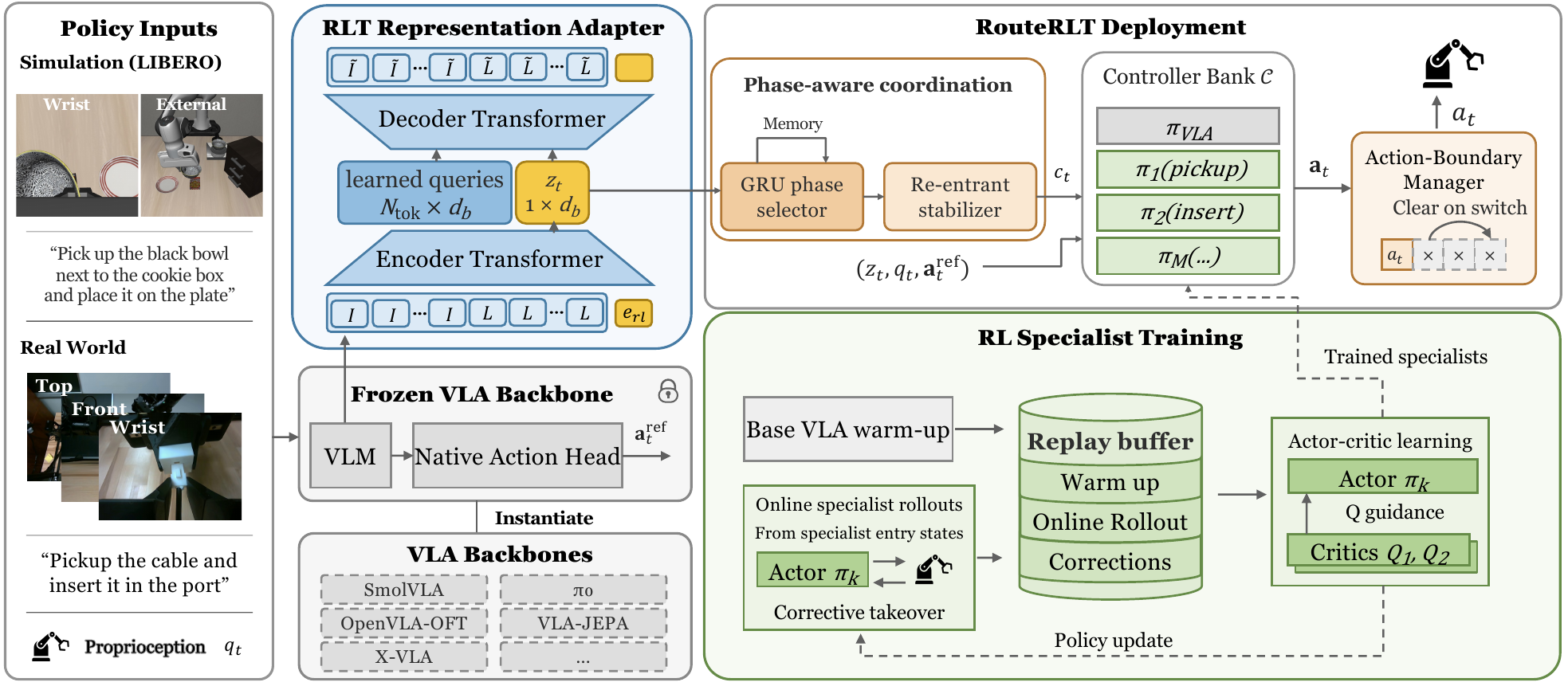}
\caption{\method{} augments a frozen VLA with phase-specific RL specialists.
A VLA-specific adapter provides compact states, while the VLA's proposed action chunks serve as references for specialist control. A GRU \emph{phase selector} proposes a controller, and a \emph{re-entrant stabilizer} governs handoffs, allowing repeated specialist invocation. On an accepted switch, the \emph{action-boundary manager} discards pending
actions and queries the new controller for the next executed action. Specialists learn from VLA warm-up, online rollouts, and corrective demonstrations. The interface is shared across backbones, while adapter and specialist weights
are backbone-specific. In terms of notation, a bold $\mathbf a_t$ denotes an action chunk, and a plain $a_t$ denotes one executed action.}
\label{fig:routerlt}
\end{figure*}

\section{Introduction}

Pretrained vision--language--action (VLA) models combine visual perception, natural language understanding, and action generation to enable general-purpose control in robotics~\cite{kim2025openvla,black2025pi0}.
While these \emph{generalist} VLAs have been shown to handle an impressively broad range of tasks, they often struggle during the precision-critical phases of multi-stage manipulation tasks. For example, contact-rich alignment in connector insertion tasks, which requires fine-grained corrective motions based on contact feedback to achieve precise alignment, remains challenging for generalist VLAs~\cite{zhu2026wirecraftsimulationbenchmarkindustrial}.

To mitigate this issue, VLA models can be used in
conjunction with reinforcement learning (RL) \cite{Sutton2018-eh} \emph{specialists} that are specifically trained to handle the precision-critical phases. For example, the RL Token (RLT) framework ~\cite{xu2026rlt} was recently proposed to address this limitation. At a high level, RLT leverages online RL to refine the behavior of pretrained VLAs on specialty tasks. RLT learns a lightweight controller (i.e., a specialist) that reads a compact summary of a (frozen) VLA's internal states and stays anchored to the action chunk that the VLA proposes. Importantly, RLT does not change the base VLA policy, such that the generalist functionality is preserved outside of the specialty task.

Although the combination of a generalist VLA and RL specialists can be powerful, the \emph{coordination} between these controllers, which dictates when a specialist should take over from the base VLA and vice-versa, is challenging and remains an open research question. For example, a specialist may leave the robot in a state that is unsuitable for a subsequent specialist~\cite{liu2026foresightresidualrllonghorizon}.
Similarly, a transition between controllers can introduce distribution shifts for the incoming controller~\cite{xin2026rpg}. Action chunking also adds a separate timing problem: fixed execution
horizons can carry stale actions into critical
phases~\cite{xu2026bcp}. This motivates the need for a coordination mechanism that: 1) determines when and which controller, the generalist or one of the specialists, should be used, and 2) executes the transition appropriately at deployment.

In this paper, we address this gap by introducing \emph{\method{}}, a modular framework that uses a learned \emph{phase-aware coordination}
mechanism to manage handoffs between a generalist VLA serving as the
default controller and RLT-based RL specialists; see
Fig.~\ref{fig:routerlt_concept}. Its interface and training procedure are applicable across VLA backbones, such that separate specialist weights can be learned for each. More formally, our contributions are as follows:
\begin{itemize}
  \item We introduce a modular refinement framework that builds on generalist VLA capabilities through the reuse and composition of RL specialists for precision-critical phases in multi-object and multi-phase manipulation;

  \item We develop a phase-aware coordination mechanism that uses causal compact-state history to select controllers, stabilize transitions, and execute immediate handoffs;

  \item We show that RouteRLT improves task completion rates, maintains these gains under perturbations, and enables effective autonomous handoffs in a physical connector pickup and insertion task.
\end{itemize}

\section{Related Work}

\subsection{Vision--Language--Action Models and Their Refinement}

Vision--Language--Action Models pursue generalization via
multimodal pretraining, whether by learning from large robot
datasets~\cite{kim2025openvla}, or by coupling a vision--language model to
a continuous action expert~\cite{black2025pi0}. Recent work has explored
different ways to improve the efficiency, adaptability, and
capabilities of these models. SmolVLA adopts a compact
architecture for efficient deployment on resource-constrained
robotic systems~\cite{shukor2025smolvla}, while OpenVLA-OFT improves fine-tuning
through parallel decoding, continuous actions, and
action chunking~\cite{kim2025oft}. Other approaches augment VLAs with
additional learning components, such as latent world-model
learning, as in VLA-JEPA~\cite{sun2026vlajepa}, or a lightweight policy module
for action generation, as in VLA-Adapter~\cite{wang2026vlaadapter}.

While VLAs have been shown to handle an impressively
broad range of tasks, they often struggle during the precision-critical
phases of multi-stage manipulation. For example, the
recent WireCraft benchmark~\cite{zhu2026wirecraftsimulationbenchmarkindustrial} shows that VLA policies
struggle with contact-rich alignment in connector insertion
tasks. Reinforcement learning (RL) provides one approach for
improving policy performance in such phases. For example,
RLDG uses task-specific RL to generate data for fine-tuning
a generalist policy~\cite{XuC1-RSS-25}, while HIL-SERL incorporates human
corrections into RL training from visual observations~\cite{luo2025hilserl}.
ConRFT adapts VLA policies through a consistency-based
policy optimization procedure~\cite{chen2025conrft}, and STARE-VLA incorporates
stage-aligned reinforcement signals during fine-tuning~\cite{xu2025starevla}.
Other approaches avoid directly modifying the
policy: DSRL steers a frozen diffusion policy through its
latent noise~\cite{wagenmaker2025dsrl}, while V-GPS selects among a generalist
policy's candidate actions using a learned value function~\cite{nakamoto2025vgps}.

Most relevant to this work is RL Token (RLT)~\cite{xu2026rlt}. RLT
compresses internal VLA features into a readout token, and
learns a lightweight actor--critic head anchored to reference
action chunks. In this work, we retain the representation
and control mechanism from RLT to train our specialists.
Also relevant is Foresight Residual RL, which addresses a
similar compositional challenge to the one considered in
this work by rewarding terminal states according to their
contribution to downstream task success~\cite{liu2026foresightresidualrllonghorizon}.
By contrast, RouteRLT retains sparse, skill-specific rewards, and addresses
composition through controller selection and handoff.

\subsection{Generalist--Specialist Control}

When various controllers operate within a shared task, a
mechanism is needed to decide which of them acts. Several
types of approaches have been proposed to tackle this
challenge. One type of approach, called interactive imitation
learning, studies when a learned policy should yield to an
expert. For example, SafeDAgger triggers on predicted action
discrepancy~\cite{zhang2017safedagger}, and LazyDAgger reduces switching through
asymmetric entry and exit conditions~\cite{hoque2021lazydagger}, a design we
retain in our framework. Related approaches use human
intervention~\cite{liu2025sirius}, detect policy failures~\cite{agia2025sentinel}, or fuse experts
instead of selecting one controller~\cite{xin2026rpg,yuan2022mes}.

A complementary line of work addresses which controller
should be active during task execution, with selection guided
by safety requirements or task progress. Recovery RL, for
example, switches between task and recovery behaviors
using a safety critic~\cite{thananjeyan2021recoveryrl}. Sparse Diffusion Policy and
AtomicVLA explore reusable expert modules and atomic
skill composition~\cite{wang2025sdp,zhang2026atomicvla}. MuST learns skill heads and
a progress estimator by imitation, using progress to skip
or repeat skills~\cite{gao2025must}. MoIRA coordinates decoupled VLA
experts using task and expert descriptions~\cite{kuzmenko2026moira}. TOWN-VLA
controls whether retrieved text may modify the prompt of
a frozen VLA, intervening at the policy input rather than
replacing the controller's robot actions~\cite{zhou2026townvla}. By contrast, in
RouteRLT, a frozen generalist remains the default controller,
while independently trained, phase-specific RL specialists can
take exclusive control, relinquish it, and be invoked again
later within the same task.

A separate question is when an ownership decision takes
effect relative to the controller's action horizon. PACE selects
replanning boundaries from the predicted action chunk's
kinematic structure~\cite{nie2026pace}, while BCP learns continue-or-replan
decisions for a frozen base policy using RL~\cite{xu2026bcp}; both make
decisions within the action horizon while retaining the same
underlying controller. PAMAE replaces a flow-matching
action expert with a phase-supervised expert mixture~\cite{yang2026pamae},
while TRACT assigns current- and next-phase authority
within a predicted horizon using a task-local graph~\cite{liu2026tract}. In
RouteRLT, routing decisions are decoupled from controller
execution horizons: an accepted ownership change can take
effect before the current chunk finishes, causing the outgoing
controller's queued actions to be discarded and the incoming
controller to generate an updated action chunk.

\section{\method{}}
\label{sec:method}

In this section, we present our primary contribution: \emph{\method{}}.
\method{} uses a frozen generalist VLA as the default controller, along with phase-specific
RL specialists that intervene where precision is required and hand control back when
the phase ends. The specialists are inspired by the compact readout and
reference-anchored control introduced by RLT~\cite{xu2026rlt}, extended here to
heterogeneous backbones and to repeated intervention within one task, across
multi-stage and multi-object task suites.

At environment step $t$ the robot observes camera images $I_t$, proprioception
$q_t$, and a task instruction $\ell$. A controller bank
$\mathcal C=\{\pi_{\mathrm{VLA}}\}\cup\Pi_{\mathrm{RL}}$ holds the frozen
generalist and the specialist suite $\Pi_{\mathrm{RL}}=\{\pi_1,\ldots,\pi_M\}$.
A task is divided into phases $\mathcal K$, and a fixed map
$g\colon\mathcal K\to\mathcal C$ assigns each critical phase to its specialist
and every remaining phase to $\pi_{\mathrm{VLA}}$. At configured query times,
\method{} selects the controller $c_t\in\mathcal C$ that owns the action
sequence sent to the robot. Fig.~\ref{fig:routerlt} shows the complete system.
The adapter is trained against the frozen VLA and then frozen for both
specialist learning and phase selection. The phase selector is trained on labeled Base rollouts and can
be further adapted using labeled specialist rollouts.
The adapter, specialists, and phase selector are frozen for evaluation and
use the same compact-state representation as during training.

\subsection{Backbone Interface}
\label{sec:adapter}

Each backbone uses the same adapter construction, which leaves its weights
frozen and exposes
\begin{equation}
\begin{aligned}
 z_t &= E(I_t,\ell)\in\mathbb R^{d_b},\\
 \tilde{\mathbf a}_t &= A(I_t,q_t,\ell)\in\mathbb R^{H\times d_a},
\end{aligned}
\label{eq:backbone_adapter}
\end{equation}
where $A$ is the backbone's action-generation path and $E$ is a readout encoder
added on top of its frozen hidden states. The compact state $z_t$ takes the
backbone's token width $d_b$, and the proposal spans that backbone's horizon
$H$ with action dimension $d_a$. The adapter chooses none of these; it
preserves the backbone's action normalization and postprocessing conventions.

$E$ uses a learned readout query $e_{\mathrm{rl}}$ to attend to the selected
frozen image and language hidden states, producing the compact state $z_t$.
Our tasks span several objects and several stages, and the instruction selects the target, so
the language states are read alongside the image states. During training a
decoder reconstructs those states from $z_t$ and $N_{\mathrm{tok}}$ learned
positional queries of width $d_b$, one per target position, so $z_t$ must
retain enough of the backbone's representation to support reconstruction.
The decoder is discarded after training.

\subsection{Phase-Specific RL Specialists}
\label{sec:specialists}

Each specialist produces a chunk of $C\le H$ actions. This length sets the
actor output, reference slice, critic input, and maximum replay-window length.
The generalist queues $C_{\mathrm{VLA}}\le H$ actions from its native prediction.
The two controllers may use different execution horizons, so
$C_{\mathrm{VLA}}$ need not equal $C$. Either chunk can be interrupted by an
accepted handoff, which replaces the pending actions with a fresh chunk
from the incoming controller. A
specialist is shared across tasks assigned to the same skill on one backbone, while
adapter and specialist weights are trained separately for each backbone.

\paragraph{Reference-anchored policy}
Specialist $\pi_k$ is conditioned on the state $s_t=(z_t,q_t)$ and a
reference chunk $\mathbf a_t^{\mathrm{ref}}$, supplied by the VLA as
$\tilde{\mathbf a}_{t,1:C}$ during autonomous execution.
$\mu_t^k$ is the mean action chunk. The specialist executes
\begin{equation}
 \hat{\mathbf a}_t^k=\mu_t^k+\epsilon,\qquad
 \epsilon\sim\mathcal N(0,\sigma^2 I),
\label{eq:specialist}
\end{equation}
with a fixed exploration scale $\sigma$ during training.
Evaluation and the actor objective use the noise-free mean $\mu_t^k$.

Corrective takeovers within a specialist's skill interval are stored
in replay with the demonstrated actions actually executed. During learning, these actions replace the corresponding
elements of $\mathbf a_t^{\mathrm{ref}}$ at expert-controlled timesteps;
the remaining elements retain the VLA proposal. The resulting reference is
used for both actor conditioning and the proximity penalty. Reference dropout
is applied to the actor input during training, while the penalty uses the
undropped reference.

The actor trades predicted return against proximity to the reference, with an
optional temporal-smoothness regularizer:
\begin{equation}
\begin{aligned}
 \mathcal L_{\mathrm{actor}}
 ={}&-\lambda_Q(n)Q_1(s_t,\mu_t^k)\\
 &+\frac{\lambda_{\mathrm{ref}}(n)}{Cd_a}
       \|\mu_t^k-\mathbf a_t^{\mathrm{ref}}\|_2^2\\
 &+\lambda_\Delta\mathcal L_\Delta(\mu_t^k),
\end{aligned}
\label{eq:actor_loss}
\end{equation}
\begin{equation}
 \mathcal L_\Delta(\mu_t^k)=
 \frac{1}{(C-1)d_a}\sum_{j=1}^{C-1}
 \|\mu_{t,j+1}^k-\mu_{t,j}^k\|_2^2 .
\label{eq:smoothness}
\end{equation}
Over training updates $n$, $\lambda_{\mathrm{ref}}$ decreases and $\lambda_Q$ increases, shifting
emphasis from reference proximity toward predicted return. The critics take the state
and an action chunk with no separate reference input, so the reference acts
through the actor's conditioning and its proximity penalty.

\paragraph{Chunk-level learning}
Twin critics with Polyak-averaged targets follow standard off-policy
practice~\cite{fujimoto2018td3}; what changes is the unit of a transition. One
transition starts at $t$ and covers up to $C$ executed actions. Its length
$L\le C$ is shortened if the skill ends, ownership changes
(Sec.~\ref{sec:boundary}), or the step limit is reached. Its target is
\begin{equation}
 y_t=\sum_{j=0}^{L-1}\gamma^j r_{t+j}
     +(1-d_t^k)\,\gamma^L\min_{i\in\{1,2\}}\bar Q_i(s_{t+L},\mathbf a'),
\label{eq:chunk_target}
\end{equation}
with $\mathbf a'$ the actor mean at $s_{t+L}$ under clipped noise. The reward
is local to the skill, one on success inside the window and zero otherwise, and
$d_t^k=1$ when the skill interval ends, including a transfer to
another controller, whether or not the skill succeeds.
A step-budget truncation instead leaves $d_t^k=0$, as do windows
that end while the skill interval continues.
Replay windows do not cross controller handoffs, but may contain
expert corrections within the same skill interval. Overlapping windows reuse
collected transitions without altering the executed trajectory, and the
reference-refresh rate inside a window follows each backbone's chunk-execution
convention.

The replay buffer opens with generalist-controlled rollouts through the skill
window and then collects rollouts from the actor's exploration distribution. A
corrective takeover is recorded when a rollout stalls, and corrections are used
in training only.

\subsection{Causal Phase Selection}
\label{sec:router}

The phase selector uses only compact-state history. Its query interval
is configured independently of the controllers' execution horizons.
Querying more frequently than either controller replans allows the boundary
manager of Sec.~\ref{sec:boundary} to act on a phase change before the current
chunk finishes. At a query time $t$,
write $t^-$ for the preceding query time; the owning controller does not
change between queries. Compact states are extracted at every environment
step; the configured interval controls selector updates, not feature extraction.

The selector is a gated recurrent unit (GRU). A projection $P_\phi$ maps each
compact state into the recurrence, which carries a hidden state forward:
\begin{equation}
 h_t=\operatorname{GRU}_\phi(P_\phi(z_t),h_{t^-}),\qquad
 \mathbf p_t=\operatorname{softmax}(H_\phi(h_t)).
\label{eq:router}
\end{equation}
The hidden state carries the compact-state history and is reset only at the
start of an episode. A fixed-history selector that concatenates a bounded window
of compact states instead of carrying a hidden state is evaluated in
Sec.~\ref{sec:ablations}.

Phase annotations are mapped through $g$ before training, so the selector's
classes are the controllers in use: the generalist and each enabled specialist.
It is trained by cross-entropy weighted inversely to training-label frequency,
so the critical phases are not swamped by the dominant generalist class. At
deployment it emits its highest-scoring class.

\subsection{Re-entrant Stabilization and Chunk Boundaries}
\label{sec:boundary}

\paragraph{Stabilizer}
Entry and exit are asymmetric, as in interactive
handoff~\cite{hoque2021lazydagger}. Let $\hat c_t$ be the controller the
selector names at query $t$, $n_t$ the number of consecutive queries that have
named a controller other than the current owner, and $u_t$ the environment steps since the current specialist took control. While the generalist owns the step,
$\hat c_t$ takes control at once. While a specialist owns it, leaving takes
$d_{\mathrm{exit}}$ consecutive queries against it, unless it has already held
for $T_{\mathrm{hold}}$ steps:
\begin{equation}
 c_t=\begin{cases}
  \hat c_t, & c_{t^-}=\pi_{\mathrm{VLA}},\\
  \hat c_t, & \hat c_t\neq c_{t^-}\ \text{and}\
             (n_t\geq d_{\mathrm{exit}}\ \text{or}\ u_t\geq T_{\mathrm{hold}}),\\
  c_{t^-}, & \text{otherwise}.
 \end{cases}
\label{eq:stabilizer}
\end{equation}
Consecutive queries against the owner need not name the same replacement, and
that replacement may be the generalist or another specialist. The lease waives
the confirmation; it does not itself end the control of a specialist the
selector keeps naming. The rule applies to every specialist, so ownership can
pass from one specialist to another directly, with no step assigned to the
generalist in between. On a switch, $n_t$ and $u_t$ reset. No controller is
locked out once it has exited, so a specialist can be entered again later in the
same task. Re-entry does not by itself detect or repair an incorrect
intervention.

\begin{algorithm}[t]
\caption{Boundary-aware \method{} deployment}
\label{alg:routerlt}
\small
\begin{algorithmic}[1]
\Require Frozen VLA, adapter $E$, specialist bank, phase selector,
instruction $\ell$, selector query interval, stabilizer parameters
$d_{\mathrm{exit}}, T_{\mathrm{hold}}$, execution horizons $C_{\mathrm{VLA}}, C$
\State $c\gets\pi_{\mathrm{VLA}}$; $\mathcal Q\gets\emptyset$;
initialize selector and stabilizer states
\For{each environment step $t$ until termination}
  \State Observe $(I_t,q_t)$; compute $z_t\gets E(I_t,\ell)$
  \If{$t$ is a query time}
    \State Advance the selector on $z_t$ and let it name $\hat c_t$
    \State Update the owner $c$ from $\hat c_t$ by \eqref{eq:stabilizer};
    clear $\mathcal Q$ if the owner changes
  \EndIf
  \If{$\mathcal Q=\emptyset$}
    \State Query the VLA for $\tilde{\mathbf a}_t$;
    set $\mathbf a_t^{\mathrm{ref}}\gets\tilde{\mathbf a}_{t,1:C}$
    \State $\mathcal Q\gets\tilde{\mathbf a}_{t,1:C_{\mathrm{VLA}}}$ if $c=\pi_{\mathrm{VLA}}$,
    else the owning specialist's chunk given $(z_t,q_t,\mathbf a_t^{\mathrm{ref}})$
  \EndIf
  \State Execute $a_t\gets\operatorname{popfront}(\mathcal Q)$
\EndFor
\end{algorithmic}
\end{algorithm}

\paragraph{Action-boundary manager}
A routing decision and a chunk boundary rarely coincide. We use bold
$\mathbf a_t$ for a selected chunk and plain $a_t$ for one executed action;
$\mathbf a_t$ is $\tilde{\mathbf a}_{t,1:C_{\mathrm{VLA}}}$ when the generalist owns the step
and $\hat{\mathbf a}_t^k$ when specialist $k$ does. Let $\mathcal Q$ hold the
current owner's remaining actions. If $c_t\neq c_{t^-}$, the manager discards
every queued action, outgoing and incoming alike, queries the new owner, and
executes the first action of the chunk it returns; a specialist query takes a
fresh reference from the same observation. If ownership is unchanged,
$\mathcal Q$ is consumed until it empties. No controller executes a chunk
planned from an observation that the switch has made stale, and the switch takes
effect at the query that decides it rather than at the next chunk boundary.
Algorithm~\ref{alg:routerlt} summarizes deployment.

\section{Experiments}
\label{sec:experiments}

\begin{figure}[H]
  \centering
  \includegraphics[width=\columnwidth]{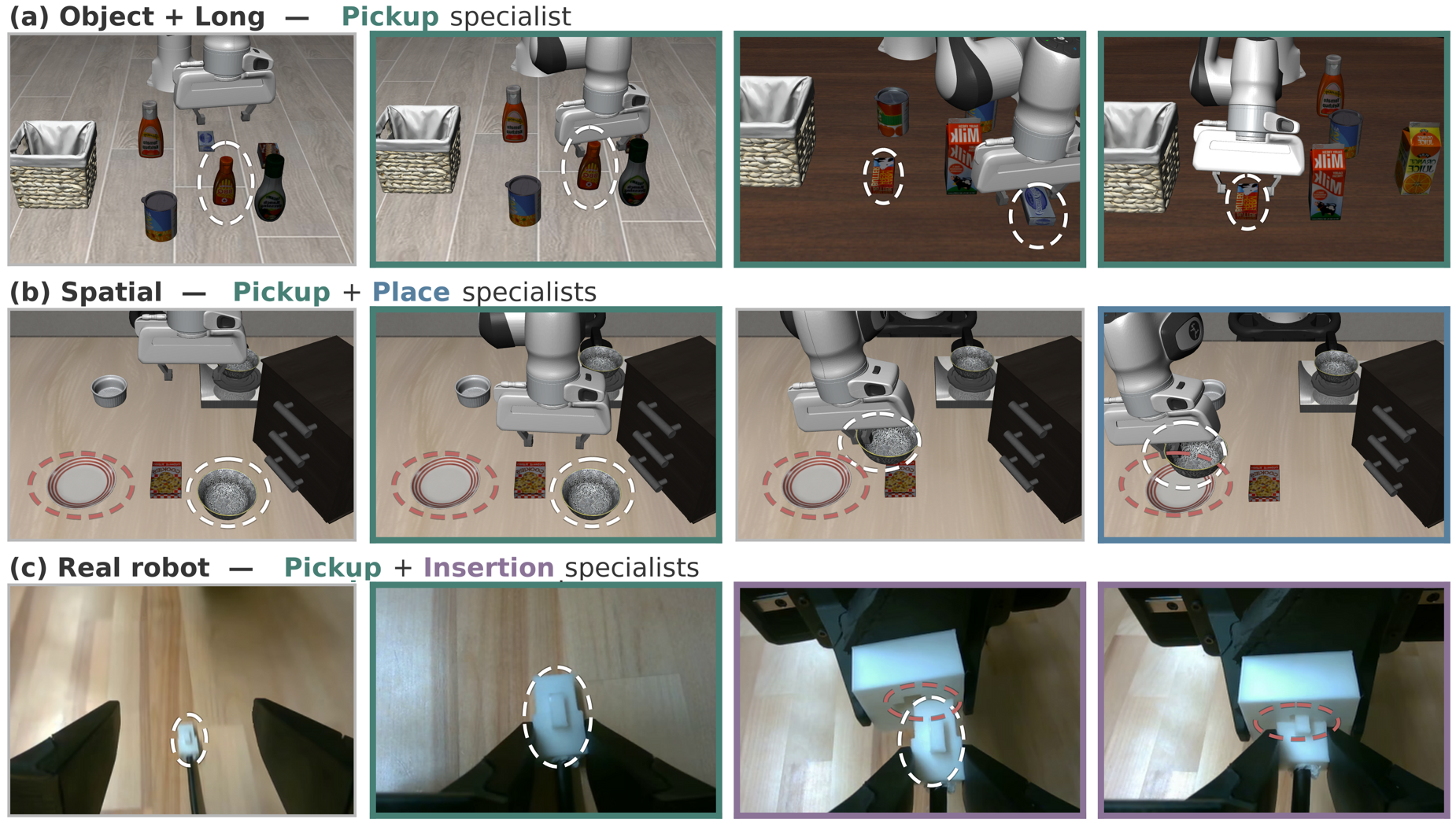}
  \caption{Illustrative specialist phases.
    (a) Object (left 2) and Long (right 2) share 1 Pickup specialist.
    (b) Spatial uses Pickup and Place specialists.
    (c) Real-robot pickup and insertion.
    Colored borders indicate specialist phases; gray indicates generalist phases.}
  \label{fig:task_overview}
\end{figure}

We evaluate whether specialists can be reused across objects and task
phases, as illustrated in Fig.~\ref{fig:task_overview}, and whether control
selection from compact-state history improves full-task success.
SmolVLA is the main testbed, with 4 additional VLA backbones assessing
the applicability of Pickup refinement. LIBERO-Plus,
component ablations, and physical experiments examine performance under
perturbations, the design choices, and autonomous execution, respectively.

\subsection{Experimental Setup}
\label{sec:simulation}

\paragraph{Task selection}
We consider 3 simulation settings.
First, LIBERO~\cite{liu2023libero} Object and a Long subset
(indices 0, 1, and 7) involve picking up objects and placing them in a
basket. Object varies the target object, while the selected Long tasks
require handling 2 objects, testing the reuse of 1 Pickup specialist
across objects and within a task.
We evaluate each method from 10 initial states per task, giving
100 Object episodes and 30 Long episodes.
Second, LIBERO-Plus~\cite{fei2026liberoplus} Object and the selected Long
tasks evaluate robustness to 7 families of out-of-distribution
perturbations using the same policies without retraining.
Each method is evaluated on 2,357 Object episodes and 570 Long episodes.
Third, LIBERO Spatial and LIBERO-Plus Spatial require picking up a bowl
and placing it on a plate, allowing us to evaluate the composition of
separately trained Pickup and Place specialists.
Each method is evaluated on 100 Spatial episodes and 2,274
LIBERO-Plus Spatial episodes.

\paragraph{Implementation and training}
\label{sec:training_details}
We start from VLAs fine-tuned on task demonstrations and freeze their
weights. For each experimental setting, Base VLA rollouts from
demonstration initial states provide data for training the compact-state
encoder and phase selector, and replay transitions for specialist
warm-up. During online training, scripted experts provide corrective
actions when progress stalls.
The specialist actor uses an MLP with 2 hidden layers of 256 units.
Each backbone has its own encoder, phase selector, and specialists.

For all VLA backbones in simulation, we train the compact-state encoder for 1 epoch
and the GRU for 30 epochs. Specialist training begins with 5k warm-up
updates, followed by online RL.
In simulation, the reference-proximity and Q weights in
Eq.~\ref{eq:actor_loss} follow
\begin{equation}
\begin{gathered}
 \alpha(n)=\operatorname{clip}\left(\frac{n-20000}{50000},0,1\right),\\
 \lambda_{\mathrm{ref}}=7-4.5\alpha(n),\quad
 \lambda_Q=0.05+0.40\alpha(n).
\end{gathered}
\label{eq:bc_q_schedule}
\end{equation}
where $n$ counts learner updates from the start of warm-up.
The GRU is queried at every environment step during simulation
evaluation.

Specialist checkpoints are selected by full-task success on LIBERO
Standard. Object and Long share a checkpoint selected by their combined
success count, while Spatial uses separately selected Pickup and Place
checkpoints. The selected checkpoints are also used for LIBERO-Plus
evaluation without further training.

\paragraph{Evaluation and metrics}
Each baseline VLA and its \method{} counterpart are evaluated under the
same simulation conditions on matched episodes.
Success rate (SR) is the percentage of episodes completed successfully.
Recoveries (Rec.) count Base failures that \method{} completes, while
regressions (Reg.) count Base successes it loses.
The change in steps to success ($\Delta$S2S) is the mean step count
for \method{} minus that for Base, measured on the same episodes
completed by both policies; negative values indicate fewer steps.
When combining suites, SR and $\Delta$S2S are weighted by the number of
evaluation episodes in each suite.

\subsection{Pickup Refinement with Different VLA Backbones}
\label{sec:backbone_extension}
\label{sec:sim_results}

Developed first on SmolVLA, \method{} improves success on both Object
and Long while reducing steps on common successes, as shown in
Table~\ref{tab:backbone_results}. These gains support reusing 1
Pickup specialist across objects and for repeated pickups within a task.
We apply the same refinement recipe to 4 additional backbones
without further tuning.

For these backbones, Object success increases slightly or remains
unchanged from already high baselines of $97$--$100\%$.
On Long, VLA-JEPA and OFT$_m$ maintain their success rates, while
X-VLA~\cite{zheng2026x} and $\pi_0$ each lose 1 success.
The gains thus do not extend uniformly beyond SmolVLA.
High Object success limits the available improvement, while the
unchanged training recipe and scripted corrections may leave other
backbones' failure modes insufficiently addressed.
Overall, the results demonstrate that the modular framework supports
different VLA backbones, although effective refinement still depends
on the backbone and task.
\begin{table}[t]
\centering
\caption{\method{} with different VLA backbones and a single Pickup
specialist per backbone on LIBERO Object and Long subset}
\label{tab:backbone_results}
\small
\setlength{\tabcolsep}{2pt}
\begin{tabular*}{\columnwidth}{@{\extracolsep{\fill}}llrrrr@{}}
\toprule
Backbone & Suite & Base \textit{SR} & \method{} \textit{SR} & $\Delta$\textit{S2S} & \textit{Rec./Reg.}\\
\midrule
SmolVLA & Object & 93.00 & \textbf{98.00} & $-1.51$ & \textbf{5}/0\\
$960/10$ & Long & 56.67 & \textbf{70.00} & $-10.38$ & \textbf{5}/1\\
\midrule
X-VLA & Object & 99.00 & \textbf{100.00} & $+4.23$ & \textbf{1}/0\\
$1024/30$ & Long & \textbf{86.67} & 83.33 & $+10.83$ & 2/3\\
\midrule
$\pi_0$ & Object & 97.00 & \textbf{99.00} & $-0.40$ & \textbf{2}/0\\
$2048/5$ & Long & \textbf{93.33} & 90.00 & $+5.44$ & 2/3\\
\midrule
VLA-JEPA & Object & \textbf{100.00} & \textbf{100.00} & $-1.56$ & 0/0\\
$2048/7$ & Long & \textbf{100.00} & \textbf{100.00} & $+1.24$ & 0/0\\
\midrule
OFT$_m$ & Object & 97.00 & \textbf{99.00} & $+0.38$ & \textbf{2}/0\\
$4096/8$ & Long & \textbf{96.67} & \textbf{96.67} & $-0.38$ & 0/0\\
\bottomrule
\end{tabular*}
\par\smallskip
\begin{minipage}{\columnwidth}
\footnotesize
Below each backbone, the first number is token dimension $d_b$ and the
second is execution cadence $C$.
OFT$_m$ denotes OpenVLA-OFT$_{mixture}$.
\end{minipage}
\end{table}

\subsection{Pickup Refinement under Perturbations}
\label{sec:plus_results}

Using the same SmolVLA Pickup specialist selected on LIBERO,
\method{} improves success on both LIBERO-Plus Object and Long
without retraining, as shown in Table~\ref{tab:main_results}.
Recoveries outnumber regressions in both suites, and common successes
require fewer steps. Although intervention can disrupt some Base
successes, these results show that selectively invoking a shared Pickup
specialist improves robustness to the tested perturbations over relying
on the frozen generalist alone.
\begin{table}[t]
\centering
\caption{SmolVLA-based \method{} with a single Pickup specialist on LIBERO-Plus Object and Long subset}
\label{tab:main_results}
\small\setlength{\tabcolsep}{2pt}
\begin{tabular*}{\columnwidth}{@{\extracolsep{\fill}}llrrrr@{}}
\toprule
Backbone & Suite & Base \textit{SR} & \method{} \textit{SR} & $\Delta$\textit{S2S} & \textit{Rec./Reg.}\\
\midrule
SmolVLA & Object & 44.17 & \textbf{47.94} & $-1.77$ & 226/137\\
 & Long & 21.40 & \textbf{25.44} & $-6.10$ & 41/18\\
\bottomrule
\end{tabular*}
\end{table}

\subsection{Composition of Pickup and Place Specialists}
\label{sec:spatial_composition}

In Spatial, the phase selector coordinates separately trained Pickup
and Place specialists with the frozen VLA within the same trajectory.

The combined policy improves full-task success on both LIBERO and
LIBERO-Plus Spatial, as shown in Table~\ref{tab:specialist_composition}.
On LIBERO, it preserves every Base success and recovers 3 failures.
Under LIBERO-Plus perturbations, recoveries exceed regressions by
80 episodes, raising success from $40.24\%$ to $43.76\%$.
Common successes also require fewer steps on LIBERO-Plus.
These results show that coordinating 2 specialists with the frozen
generalist improves complete-task performance, with gains retained
under the tested perturbations.
\begin{table}[t]
\centering
\caption{SmolVLA-based \method{} with Pickup and Place specialists on LIBERO and LIBERO-Plus Spatial}
\label{tab:specialist_composition}
\small\setlength{\tabcolsep}{2pt}
\begin{tabular*}{\columnwidth}{@{\extracolsep{\fill}}llrrrr@{}}
\toprule
Backbone & Suite & Base \textit{SR} & \method{} \textit{SR} & $\Delta$\textit{S2S} & \textit{Rec./Reg.}\\
\midrule
SmolVLA & LIBERO & 91.00 & \textbf{94.00} & $+1.62$ & 3/0\\
 & Plus & 40.24 & \textbf{43.76} & $-2.25$ & 172/92\\
\bottomrule
\end{tabular*}
\end{table}

\subsection{\method{} Ablations}
\label{sec:ablations}

Table~\ref{tab:simulation_ablations} examines 4 design choices:
history, routing rule, handoff timing, and representation.
Within each variant, 1 Pickup specialist serves both LIBERO Object
and the Long subset. All variants use the same evaluation episodes
and are evaluated at a training budget of 400 online episodes.
The SmolVLA Base and full-method results are shared with
Table~\ref{tab:backbone_results}.
The Object and Long columns report success counts.

\begin{table}[t]
\centering
\caption{SmolVLA-based \method{} ablations with a single Pickup specialist on LIBERO Object and Long subset}
\label{tab:simulation_ablations}
\small
\setlength{\tabcolsep}{2pt}
\begin{tabular*}{\columnwidth}{@{\extracolsep{\fill}}lrrrrr@{}}
\toprule
Variant & Object & Long & \textit{SR} $\uparrow$ & $\Delta$\textit{S2S} $\downarrow$ & \textit{Rec./Reg.}\\
\midrule
SmolVLA Base & 93 & 17 & 84.62 & -- & --\\
\textbf{\method{}} & \textbf{98} & 21 & \textbf{91.54} & $-3.55$ & \textbf{10}/1\\
\midrule
\multicolumn{6}{@{}l}{\emph{History}}\\
MLP, $W=1$ & 96 & 19 & 88.46 & $-3.21$ & 8/3\\
MLP, $W=2$ & 96 & \textbf{22} & 90.77 & $-1.03$ & \textbf{10}/2\\
MLP, $W=4$ & 97 & 21 & 90.77 & $\mathbf{-3.56}$ & 9/1\\
MLP, $W=6$ & \textbf{98} & 21 & \textbf{91.54} & $-1.59$ & 9/\textbf{0}\\
MLP, $W=8$ & 96 & 20 & 89.23 & $-1.97$ & 8/2\\
\midrule
\multicolumn{6}{@{}l}{\emph{Routing rule}}\\
Rule-window & 95 & 18 & 86.92 & $-0.78$ & 8/5\\
\midrule
\multicolumn{6}{@{}l}{\emph{Handoff timing}}\\
Chunk-boundary & 94 & 18 & 86.15 & $-3.31$ & 6/4\\
\midrule
\multicolumn{6}{@{}l}{\emph{Representation}}\\
Image-only & 93 & 18 & 85.38 & $-0.22$ & 7/6\\
\bottomrule
\end{tabular*}
\end{table}

\paragraph{History}
What the phase selector sees shapes when it intervenes and, in turn, the
state from which the specialist begins. We compare MLPs using $W$ compact states sampled 10 environment steps
apart with a GRU that carries history in its hidden state.
Success spans $88.46$--$91.54\%$, peaking at
$W=6$ before falling at $W=8$. The GRU matches the best fixed window.
Temporal context helps, but simply retaining more
of it is not enough. We use the GRU to carry that context without fixing
a window length whose suitability may change across tasks.

\paragraph{Routing rule}
Rule-window uses privileged simulator geometry to determine when
the Pickup specialist acts. It achieves $86.92\%$ success, compared
with $91.54\%$ for learned phase selection. This comparison supports
using compact-state history to determine intervention timing rather
than relying on the hand-specified geometric rule, without requiring
privileged simulator input at deployment.

\paragraph{Handoff timing}
Deferring accepted switches until the current chunk ends reduces
success to $86.15\%$. During the delay, the outgoing controller
continues to act despite the decision to transfer control.
Applying the switch at the next action avoids this delay and improves
full-task success.

\paragraph{Representation}
SmolVLA processes image and language tokens jointly through multimodal
attention. The image-only and image-plus-language variants each use their
own separately trained encoder, phase selector, and specialist. Including explicit language tokens in the compact representation
raises success from $85.38\%$ to $91.54\%$. These tokens provide useful
task context for intervention beyond what the image tokens alone supply. 

\subsection{Real-Robot Pickup and Insertion}
\label{sec:hardware}

\begin{figure}[t]
  \centering
  \setlength{\abovecaptionskip}{2pt}
  \includegraphics[width=\columnwidth,trim=0 1bp 0 0,clip]{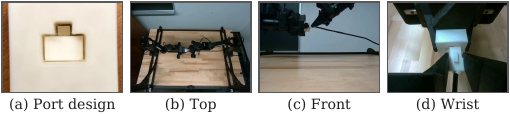}
  \caption{Hardware setup for pickup and insertion: (a) the port design with 1 mm clearance and
  (b)--(d) 3 camera views at the same time $t$.}
  \label{fig:hardware_setup}
\end{figure}

\paragraph{Hardware and task}
We evaluate connector pickup and insertion on a Trossen Stationary AI
platform~\cite{trossenstationaryai}.
The connector has an insertion tip, a graspable housing, and a latch,
with its initial position varied within a $2\times2$\,cm region.
The left arm holds the female port at a nominally fixed pose,
although contact can cause small displacements.
The policy receives top, front, and wrist camera images and controls
the right arm's joint positions and gripper opening.
See Fig.~\ref{fig:hardware_setup} for the port design and camera views.

\paragraph{Training}
SmolVLA is fine-tuned on 48 demonstrations and then frozen.
The compact-state encoder receives 20k updates with a batch size of 16.
Warm-up replay combines human demonstrations with 20 VLA rollouts
for Pickup and 50 for Insertion.
Each specialist receives 30k critic-only updates, followed by
20k critic and 10k actor updates, before online refinement.

During continuous execution, variations in Pickup produce arm
configurations and grasp poses outside the initial training
distribution, affecting phase selection and subsequent Insertion.
We therefore jointly refine both specialists and the GRU.
The operator selects controllers and corrects stalled execution,
then returns control to the policy.
Both specialists receive online RL updates from rollout replay,
while phase labels from 15 continuous trajectories support
supervised GRU adaptation.

\paragraph{Deployment and evaluation}
The VLA executes 35 actions per chunk, each specialist executes 10,
and the GRU is queried every 5 control steps. The stabilizer uses $d_{\mathrm{exit}}=3$ selector queries and
$T_{\mathrm{hold}}=30$ control steps.
Each method is evaluated over 35 trials with a 60\,s budget,
with retries allowed within a trial.
Pickup requires a stable housing grasp. Full-task success requires fully seated insertion.

\begin{figure}[t]
  \centering
  \setlength{\abovecaptionskip}{2pt}
  \includegraphics[width=\columnwidth]{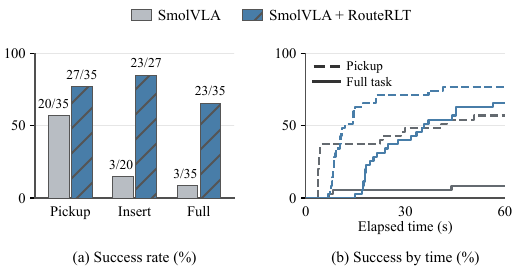}
  \caption{SmolVLA-based \method{} on physical connector pickup and insertion.
  (a) success rates, with insertion conditioned on Pickup success.
  (b) cumulative Pickup and full-task success over all 35 trials per
  method within a 60\,s budget, timed from the first action dispatch.
  Full-task event times are estimated from recorded successful endings.}
  \label{fig:hardware_results}
\end{figure}

\paragraph{Results}
As shown in Fig.~\ref{fig:hardware_results}, \method{} completes
23 of 35 trials ($65.7\%$), compared with 3 of 35 ($8.6\%$)
for SmolVLA, an improvement of 57.1 percentage points.
By 30\,s, \method{} has completed 14 trials, compared with 2
for SmolVLA. The higher completion rates at 30\,s and at the 60\,s limit support
more reliable full-task execution through local specialist refinement
and autonomous handoffs.

\section{Limitations}
Our evaluation leaves several questions for further study.
LIBERO-Plus measures robustness at the full-task level, but does not
separate the effects of distribution shift on phase selection and
specialist execution. Examining these effects for individual
perturbation types would help identify where further refinement is
needed. Specialist reuse across objects is demonstrated in simulation,
while the physical experiments focus on one connector geometry.
Extending the hardware study to different objects would test this
capability beyond simulation. Finally, each VLA backbone requires its
own trained encoder, phase selector, and specialists. Under our
computational budget, we concentrate the broader suite and perturbation
evaluations on SmolVLA, leaving similarly extensive evaluations of
the other backbones for future work.

\section{Conclusion}
\method{} extends local RL refinement to manipulation involving
multiple objects and phases without updating the generalist.
The experiments demonstrate that a shared specialist can improve
tasks involving different objects and repeated skill invocation,
while distinct specialists can be composed with the VLA to improve
task completion in simulation and on a physical robot.
Evaluation with 5 VLA backbones establishes the applicability of the
modular interface, and SmolVLA retains net gains under the tested
LIBERO-Plus perturbations.
Ablations further show that compact-state history, explicit language
tokens, and timely handoffs contribute to successful intervention.
Altogether, these results establish reusable specialist learning and
phase-aware coordination as a practical approach towards improving
performance on complex, multi-stage manipulation tasks.

\section*{Acknowledgment}
The OpenAI image-generation tool, accessed through Codex, generated the
conceptual illustration in Fig.~\ref{fig:routerlt_concept}.

\bibliographystyle{IEEEtran}
\bibliography{arxiv/references}

\end{document}